\documentclass[11pt,a4paper]{article}
\usepackage{authblk}
\usepackage[hyperref]{naaclhlt2019}
\usepackage{times}
\usepackage{latexsym}
\usepackage{url}
\usepackage{amsmath}
\usepackage{amsfonts}
\usepackage{booktabs}
\usepackage{graphicx}
\usepackage{todonotes}
\usepackage{float}
\usepackage{enumitem}
\usepackage{caption}

\aclfinalcopy 

\title{Improving Lemmatization of Non-Standard Languages with Joint Learning}

\author[1]{\textbf{Enrique Manjavacas}}
\author[2]{\textbf{{\'A}kos K{\'a}d{\'a}r}}
\author[1]{\textbf{Mike Kestemont}}
\affil[1]{University of Antwerp\\CLiPS\\\tt\{firstname,lastname\}@uantwerpen.be}
\affil[2]{Tilburg University\\Tilburg Center for Cognition and Communication\\\tt{a.kadar@uvt.nl}}

\date{}

\begin{document}

\maketitle

\begin{abstract}
Lemmatization of standard languages is concerned with (i) abstracting over morphological differences and (ii) resolving token-lemma ambiguities of inflected words in order to map them to a dictionary headword. In the present paper we aim to improve lemmatization performance on a set of non-standard historical languages in which the difficulty is increased by an additional aspect (iii): spelling variation due to lacking orthographic standards.
We approach lemmatization as a string-transduction task with an encoder-decoder architecture which we enrich with sentence context information using a hierarchical sentence encoder. We show significant improvements over the state-of-the-art when training the sentence encoder jointly for lemmatization and language modeling. Crucially, our architecture does not require POS or morphological annotations, which are not always available for historical corpora. Additionally, we also test the proposed model on a set of typologically diverse standard languages showing results on par or better than a model without enhanced sentence representations and previous state-of-the-art systems.
Finally, to encourage future work on processing of non-standard varieties, we release the dataset of non-standard languages underlying the present study, based on openly accessible sources.


  
\end{abstract}

\section{Introduction}\label{sec:intro}
Lemmatization is the task of mapping a token to its corresponding dictionary head-form to allow downstream applications to abstract away from orthographic and inflectional variation \cite{lemma}. While lemmatization is considered to be solved for analytic and resource-rich languages such as English, it remains an open challenge for morphologically complex (e.g. Estonian, Latvian) and low-resource languages with unstable orthography (e.g. historical languages). Especially for languages with higher surface variation, lemmatization plays a crucial role as a preprocessing step for downstream tasks such as topic modeling, stylometry and information retrieval.

In the case of standard languages, lemmatization complexity arises primarily from two sources: (i) morphological complexity affecting the number of inflectional patterns a lemmatizer has to model and (ii) token-lemma ambiguities (e.g. ``living" can refer to lemmas ``living" or ``live") which require modeling sentence context information. In the case of historical languages, however, the aforementioned spelling variation introduces further complications. For instance, the regularity of the morphological system is drastically reduced since the evidence supporting token-lemma mappings becomes more sparse. As an example, while the modern Dutch lemma ``jaar" (en. ``year") can be inflected in 2 different ways (``jaar", ``jaren"), in a Middle Dutch corpus used in this study it is found in combination with 70 different forms (``iare", ``ior", ``jaer", etc.). Moreover, spelling variation increases token-lemma ambiguities by conflating surface realizations of otherwise unambiguous tokens---e.g. Middle Low German ``bath" can refer to lemmas ``bat" (en. ``bad") and ``bidden" (en. ``bet") due to different spellings of the dental occlusive in final position. 

Spelling variation is not exclusive of historical languages and it can be found in contemporary forms of on communication, such as micro-blogs, with loose orthographic conventions \cite{crystal}. An important difference, however, is that while for modern languages normalization is feasible \cite{schulz2016multimodular}, for many historic languages such is not possible, because one is dealing with an amalgam of regional dialects that lacked any sort of supra-regional variant functioning as target domain \cite{kestemont2016lemmatization}.


In the present paper, we apply representation learning to lemmatization of historical languages. Our method shows improvements over a plain encoder-decoder framework, which reportedly achieves state-of-the-art performance on lemmatization and morphological analysis \cite{Bergmanis2018ContextLematus,N18-1202}. In particular, we make the following contributions:
\begin{enumerate}
    \item We introduce a simple joint learning approach based on a bidirectional Language Model (LM) loss and achieve relative improvements in overall accuracy of 7.9\% over an encoder-decoder trained without joint loss and 30.72\% over edit-tree based approaches.
    \item We provide a detailed analysis of the linguistic and corpus characteristics that explain the amount of improvement we can expect from LM joint training.
    \item We probe the hidden representations learned with the joint loss and find them significantly better predictors of POS-tags and other morphological categories than the representations of the simple model, confirming the efficiency of the joint loss for feature extraction.
\end{enumerate}

Additionally, we test our approach on a typologically varied set of modern standard languages and find that the joint LM loss significantly improves lemmatization accuracy of ambiguous tokens over the encoder-decoder baseline (with a relative increase of 15.1\%), but that, in contrast to previous literature \cite{P17-1136,Bergmanis2018ContextLematus}, the overall performance of encoder-decoder models is not significantly higher than that of edit-tree based approaches. Taking into account the type of inflectional morphology dominating in a particular language, we show that the benefit of encoder-decoder approaches is highly dependent on typological morphology. Finally, to assure reproducibility, all corpus preprocessing pipelines and train-dev-test splits are released. With this release, we hope to encourage future work on processing of lesser studied non-standard varieties.\footnote{
    Datasets and training splits are available at \url{https://www.github.com/emanjavacas/pie-data}.
    Experiments are conducted with our framework \texttt{pie} available at: \url{https://www.github.com/emanjavacas/pie}. All our experiments are implemented using \texttt{PyTorch} \cite{paszke2017pytorch}.
}

\section{Related Work}

Modern data-driven approaches typically treat lemmatization as a classification task where classes are represented by binary edit-trees induced from the training data. Given a token-lemma pair, its binary edit-tree is induced by computing the prefix and suffix around the longest common subsequence, and recursively building a tree until no common character can be found. Such edit-trees manage to capture a large proportion of the morphological regularity, especially for languages that rely on suffixation for morphological inflection (e.g. Western European languages), for which such methods were primarily designed. 

Based on edit-tree induction, different lemmatizers have been proposed. For example, \citet{CHRUPALA08.594} use a log-linear model and a set of hand-crafted features to decode a sequence of edit-trees together with the sequence of POS-tags using a beam-search strategy. A related approach is presented by \citet{Gesmundo2012LemmatisationTask}, where edit-trees are extracted using a non-recursive version of the binary edit-tree induction approach. More recently, \citet{Cotterell2015JointLEMMING} have used an extended set of features and a second-order CRF to jointly predict POS-tags and edit-trees with state-of-the-art performance. Finally, \citet{P17-1136} employed a softmax classifier to predict edit-trees based on sentence-level features implicitly learned with a neural encoder over the input sentence. 

With the advent of current encoder-decoder architectures, lemmatization as a string-transduction task has gained interest partly inspired by the success of such architectures in Neural Machine Translation (NMT). For instance, \citet{Bergmanis2018ContextLematus} apply a state-of-the-art NMT system with the lemma as target and as source the focus token with a fixed window over neighboring tokens. Most similar to our work is the approach by \citet{kondratyuk2018lemmatag}, which conditions the decoder on sentence-level distributional features extracted from a sentence-level bidirectional RNN and morphological tags. 

Recently, work on non-standard historical varieties has focused on spelling normalization using rule-based, statistical and neural string-transduction models \cite{W14-0605,C16-1013,C18-1112}. Previous studies on lemmatization of historical variants focused on evaluating off-the-shelf systems. For instance, \citet{EGER16.656} evaluates different pre-existing models on a dataset of German and Medieval Latin, and \citet{dereza2018lemmatization} focuses on Early Irish. The most similar to the present paper in this area is work by \citet{kestemont2016lemmatization}, which tackled lemmatization of Middle Dutch with a neural encoder that extracts character and word-level features from a fixed-length token window and predicts the target lemma from a closed-set of true lemmas.

Using Language Modeling as a task to extract features in a Transfer Learning setup has gained momentum only in the last year, partly thanks to overall improvements over previous state-of-the-art across multiple tasks (NER, POS, QA, etc.). Different models have been proposed around the same idea varying in implementation, optimization and task definition. For instance, \citet{P18-1031} present a method to fine-tune a pretrained LM for text classification. \citet{N18-1202} learn task-specific weighting schemes over different layer features extracted by a pretrained bidirectional LM. Recently, \citet{C18-1139} used context-sensitive word-embeddings extracted from a bidirectional character-level LM to improve NER, POS-tagging and chunking. 


\section{Proposed Model}
Here we describe our encoder-decoder architecture for lemmatization. In Section~\ref{sec:model} we start by describing the basic formulation known from the machine translation literature. Section~\ref{sec:sentential} shows how sentential context is integrated into the decoding process as an extra source of information. Finally, Section~\ref{sec:finetune} describes how we learn richer representations for the encoder through the addition of an extra language modeling task.

\subsection{Encoder-Decoder}\label{sec:model}
We employ a character-level Encoder-Decoder architecture that takes an input token $x_t$ character-by-character and has as goal the character-level decoding of the target lemma $l_t$ conditioned on an intermediate representation of $x_t$.
For token $x_t$, a sequence of token character embeddings ${c^x_1, \ldots , c^x_n}$ is extracted from embedding matrix $W_{enc} \in \mathbb{R}^{|C| \times d}$ (where $|C|$ and $d$ represent, respectively, the size of the character vocabulary and the embedding dimensionality). These are then passed to a bidirectional RNN encoder, that computes a forward and a backward sequence of hidden states: $\overrightarrow{h_1^{enc}}, \ldots, \overrightarrow{h_n^{enc}}$ and $\overleftarrow{h_1^{enc}}, \ldots, \overleftarrow{h_n^{enc}}$. The final representation of each character $i$ is the concatenation of the forward and the backward states: $h_i^{enc} = [\overrightarrow{h_{i}^{enc}}; \overleftarrow{h_{i}^{enc}}]$.

At each decoding step $j$, a RNN decoder generates the hidden state $h_j^{dec}$, given the lemma character embedding $c^l_j$ from embedding matrix $W_{dec} \in \mathbb{R}^{|L| \times d}$, the previous hidden state  $h_{j-1}^{dec}$ and additional context. This additional context consists of a summary vector $r_j$ obtained via an attentional mechanism \cite{bahdanau2014neural} that takes as input the previous decoder state $h_{j-1}^{dec}$ and the sequence of encoder activations $h_1^{enc}, \ldots, h_n^{enc}$.\footnote{
    We refer to \citet{bahdanau2014neural} for the description of the attentional mechanism.
} Finally, the output logits for character $j$ are computed by a linear projection of the current decoder state $h_j^{enc}$ with parameters $O \in \mathbb{R}^{H \times |L|}$, which are normalized to probabilities with the $softmax$ function. The model is trained to maximize the probability of the target character sequence expressed in Equation~\ref{eq:seq2seq} using teacher-forcing.

\begin{equation}\label{eq:seq2seq}
    P(l_t|x_t) = \prod_{j=1}^{m} P(c^l_j|c^l_{<j}, r_j; \theta_{enc}, \theta_{dec})
\end{equation}

\subsection{Adding sentential context}
\label{sec:sentential}

Lemmatization of ambiguous tokens can be improved by incorporating sentence-level information. Our architecture is similar to \citet{kondratyuk2018lemmatag} in that it incorporates global sentence information by extracting distributional features with a hierarchical bidirectional RNN over the input sequence of tokens $x_1, ..., x_m$. For each token $x_t$, we first extract word-level features re-using the last hidden state of character-level bidirectional RNN Encoder from Section~\ref{sec:model} $w_t = [\overrightarrow{h_{t}^{enc}}; \overleftarrow{h_{t}^{enc}}]$. Optionally, word-level features can be enriched with extra lookup parameters from an embedding matrix $W_{word} \in \mathbb{R}^{|V| \times e}$ -- where $V$ and $e$ denote respectively the vocabulary size in words and the word embedding dimensionality.\footnote{
    During development word embeddings did not contribute significant improvements on historical languages, and we therefore exclude them from the rest of the experiments. It must be noted, however, that word embeddings might still be helpful for lemmatization of standard languages where the type-token ratio is smaller as well as when pretrained embeddings are available.
}
Given these word-level features $w_t$, the sentence-level features $s_t$ are computed as the concatenation of forward and backward activations of an additional sentence-level bidirectional RNN $s_t = [\overrightarrow{s_t}; \overleftarrow{s_t}]$.

In order to perform sentence-aware lemmatization for token $x_t$, we condition the decoder on the sentence-level encoding $s_t$ and optimize the probability given by Equation~\ref{eq:seq2seq+sentence}.

\begin{equation}
    \label{eq:seq2seq+sentence}
    P(l_t|x_t) = \prod_{j=1}^{m} P(c^l_j|c^l_{<j}, r_j, s_t; \theta_{enc}, \theta_{dec})
\end{equation}

\begin{figure}[ht]
    \centering
    \includegraphics[width=1\linewidth]{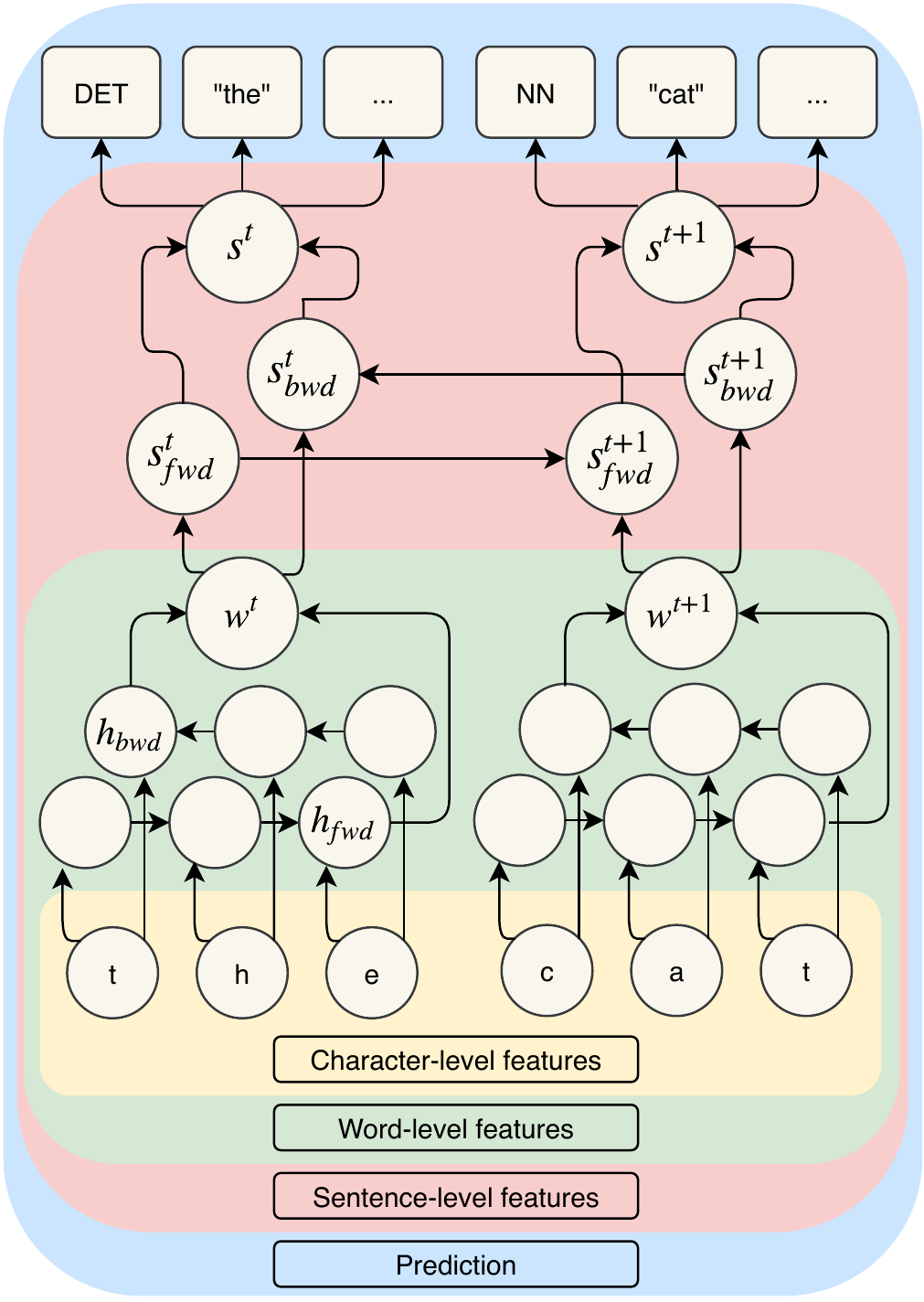}
    \caption{Hierarchical sentence encoding architecture with feature extraction at different levels.}
    \label{fig:sentence_encoder}
\end{figure}

Our architecture ensures that both word-level and character-level features of each input token in a sentence can contribute to the sentence-level features at any given step and therefore to the lemmatization of any other token in the sentence. From this perspective, our architecture is more general than those presented in \citet{kestemont2016lemmatization,Bergmanis2018ContextLematus}, where sentence information is included by running the encoder over a predetermined fixed-length window of neighboring characters. Moreover, we let the character-level embedding extractor and the lemmatizer encoder share parameters in order to amplify the training signal coming into the latter. Figure~\ref{fig:sentence_encoder} visualizes the proposed architecture.

\subsection{Improved sentence-level features} 
\label{sec:finetune}

We hypothesize that the training signal from lemmatization alone might not be enough to extract sufficiently high quality sentence-level features. As such we include an additional bidirectional word-level language-model loss over the input sentence. Given the forward and backward subvectors of the sentence encoding $s^t = [\overrightarrow{s^t}; \overleftarrow{s^t}]$, we train two additional softmax classifiers
to predict token $x^{t+1}$ given $\overrightarrow{s_t}$ and $x^{t-1}$ given $\overleftarrow{s_t}$ with parameters  $O_{LMfwd}$ and $O_{LMbwd} \in \mathbb{R}^{S \times |V|}$.\footnote{
    We have found the joint loss most effective when both forward and backward classifiers shared parameters.
}


We train our model to jointly minimize the negative log-likelihood of the probability defined by Equation~\ref{eq:seq2seq+sentence} and the LM probability defined by Equation~\ref{eq:lm}.

\begin{equation}\label{eq:lm}
    \begin{split}
    P_{LM}(\textbf{x}) & = 1/2\prod_{t=2}^{n} P(x_{t}|x_1, \ldots, x_{t-1}) \\
    & + 1/2\prod_{t=1}^{n-1} P(x_{t}|x_{t+1}, \ldots, x_n)
    \end{split}
\end{equation}

Following a Multi-Task Learning \cite{caruana1997multitask}, we set a weight on the LM negative log-likelihood which we decrease over training based on lemmatization accuracy on development data to reduce its influence on training after convergence.


\section{Experiments}

Section~\ref{sec:dataset} first introduces the datasets, both the newly introduced dataset of historical languages, and the dataset of modern standard languages sampled from Universal Dependencies (v2.2) corpus \cite{NIVRE16.348}. Finally, Section~\ref{sec:modelexp} describes model training and settings in detail.

\subsection{Datasets}\label{sec:dataset}

\begin{figure}[!ht]
    \centering
    \includegraphics[width=.95\linewidth,keepaspectratio]{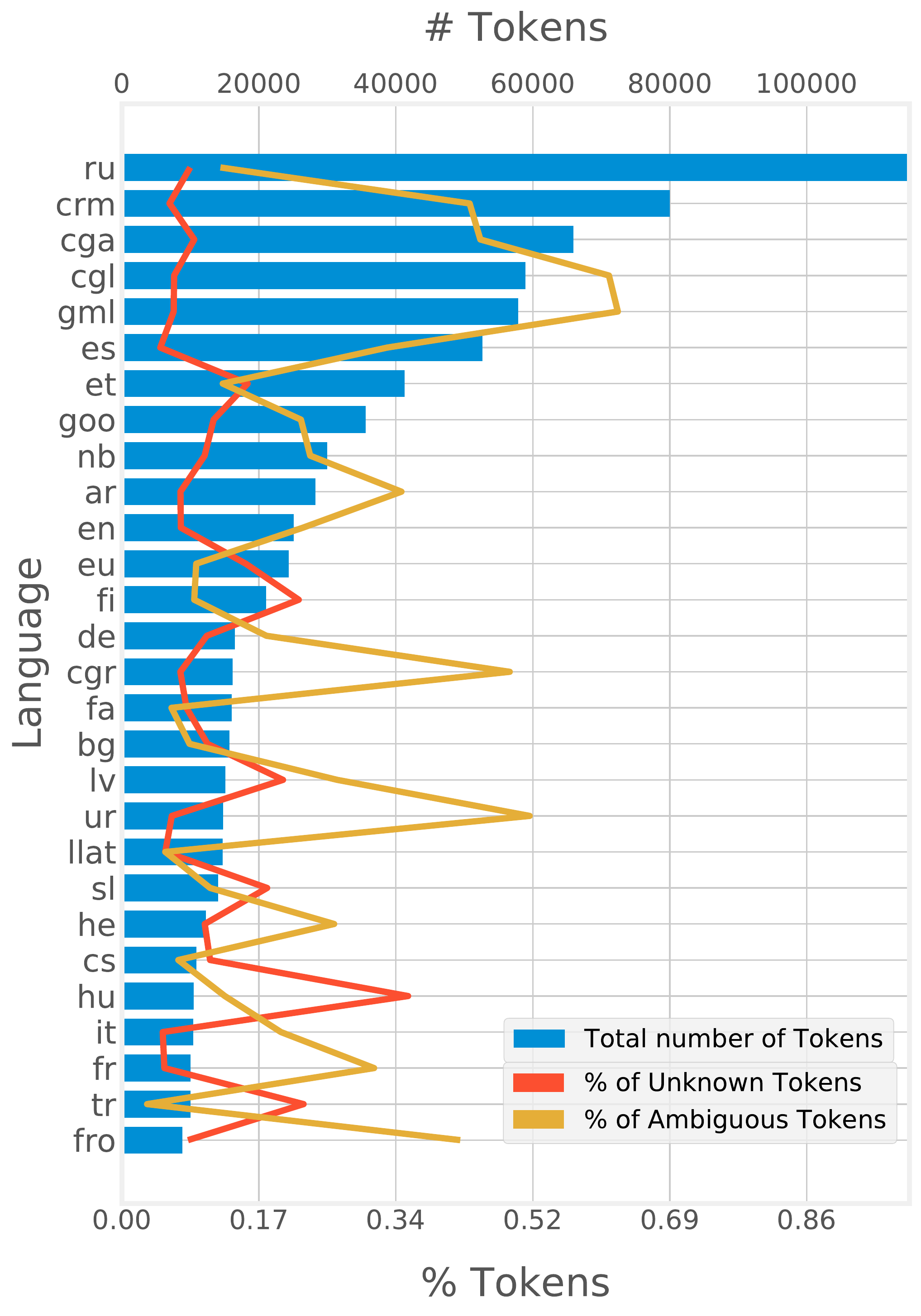}
    \caption{Statistics of total number of tokens, ambiguous and unknown tokens in the test sets. The full list of languages for both historical and standard corpora as well as the corresponding ISO 639-1 codes used in the present study can be found in the Appendix. Statistics for unknown and ambiguous tokens are shown as percentages.}
    \label{fig:test_stats}
\end{figure}

\paragraph{Historical Languages} In recent years, a number of historical corpora have appeared thanks to an increasing number of digitization initiatives \cite{piotrowski2012natural}. For the present study, we chose a representative collection of medieval and early modern datasets, favoring publicly available data, corpora with previously published results and datasets covering multiple genres and historic periods. We include a total of 8 corpora covering Middle Dutch, Middle Low German, Medieval French, Historical Slovene and Medieval Latin, which we take from the following sources.

Both \texttt{cga} and \texttt{cgl} contain medieval Dutch material from the Gysseling corpus curated by the Institute for Dutch Lexicology\footnote{
    \url{https://ivdnt.org/taalmaterialen}.
} \texttt{cga} is a charter collection (administrative documents), whereas \texttt{cgl} concerns a variety of literary texts that greatly vary in length. \texttt{crm} is another Middle Dutch charter collection from the 14th century with wide geographic coverage \cite{crm, vanHalteren2013}. \texttt{cgr}, finally, is a smaller collection of samples from Middle Dutch religious writings that include later medieval texts \cite{kestemont2016lemmatization}. \texttt{fro} offers a corpus of Old French heroic epics, known as \emph{chansons de geste} \cite{geste}. \texttt{llat} dataset is taken from the Late Latin Charter Treebank, consisting of early medieval Latin documentary texts \cite{korkiakangas2013abbreviations}. \texttt{goo} comes from the reference corpus of historical Slovene, sampled from 89 texts from the period 1584-1899 \cite{11356/1025}. \texttt{gml} refers to the reference corpus of Middle Low German and Low Rhenish texts, found in manuscripts, prints and inscriptions \cite{barteld2017referenzkorpus}. Finally, \texttt{cap} is a corpus of early medieval Latin ordinances decreed by Carolingian rulers \cite{EGER16.656}.

\paragraph{Standard Languages} For a more thorough comparison between systems across domains and a better examination of the effect of the LM loss, we evaluate our systems on a set of 20 standard languages sampled from the UD corpus, trying to guarantee typological diversity while selecting datasets with at least 20k words. We use the pre-defined splits from the original UD corpus (v2.2).\footnote{
    The full list of languages for both historical and standard corpora as well as the corresponding ISO 639-1 codes used in the present study can be found in the Appendix. In the cases where train-dev-test splits were not pre-defined, we randomly split sentences using 10\% and 5\% for test and dev respectively.
}. Figure~\ref{fig:test_stats} visualizes the test set sizes in terms of total, ambiguous and unknown tokens for both historical and standard languages.

\subsection{Models}\label{sec:modelexp}
We refer to the full model trained with joint LM loss by \texttt{Sent-LM}. In order to test the effectiveness of sentence information and the importance of enhancing the quality of the sentence-level feature extraction, we compare against a simple encoder-decoder model without sentence-level information (\texttt{Plain}) and a model trained without joint LM loss (\texttt{Sent}). Moreover, we compare to previous state-of-the-art lemmatizers based on binary edit-tree induction: \texttt{Morfette} \cite{CHRUPALA08.594} and \texttt{Lemming} \cite{Cotterell2015JointLEMMING}, which we run with default hyperparameters.


For all our models, we use the same hyperparameter values as follows. All recurrent layers have 150 cells per layer and use GRUs \cite{W14-4012}. Encoder and Decoder have 2 layers but the sentence encoder has only 1. We apply 0.25 dropout \cite{Srivastava2014} after the embedding layer and before the output layer and 0.25 variational dropout \cite{gal2016theoretically} in between recurrent layers. Models are optimized with Adam \cite{Kingma2015} using an initial learning rate of 1e-3 which is reduced by 25\% after each epoch without improvement on development accuracy. Models are trained until failing to achieve any improvement for 3 consecutive epochs. Initial LM loss weight is set to 0.2 and it is halved each epoch after two consecutive epochs without achieving any improvements on development perplexity.

We use sentence boundaries when given and otherwise use POS tags corresponding to full stops as clues. In any case, sentences are split into chunks of maximum 35 words to accommodate to limited memory. Target lemmas during both training and testing are lowercased in agreement with the implementation of \texttt{Lemming} and \texttt{Morfette}, which also do so. For models with joint loss, we truncate the output vocabulary to the top 50k most frequent words for similar reasons. We run a maximum of 100 optimization epochs in randomized batches containing 25 sentences each. The learning rate is decreased by a factor of 0.75, after every 2 epochs without accuracy increase on held-out data and learning stops after failing to improve for 5 epochs. Decoding is done with beam search with a beam size of 10.\footnote{
    For all languages, we observed relatively small gains ranging from 0.1\% to 0.5\% in overall accuracy.
}

\section{Results}\label{sec:results}

As is customary, we report exact-match accuracy on target lemmas. Besides overall accuracy, we also compute accuracy of ambiguous tokens (i.e. tokens that map to more than 1 lemma in the training data) and unknown tokens (i.e. tokens that do not appear in the training data).

\subsection{Historical languages}\label{sec:hist}
\begin{table}[!ht]
    \centering
     \resizebox{0.45\textwidth}{!}{
    \begin{tabular}{lccc}
\toprule
         & Full & Ambiguous & Unknown  \\
\midrule 
   \texttt{Edit-tree}    & 91.11          & 91.79 & 35.48  \\
   \texttt{Plain}        & 91.61          & 87.39 & 65.69  \\
   \texttt{Sent}     & 93.4           & 91.14 & \textbf{66.98}  \\
   \texttt{Sent-LM}  & \textbf{94.0}  & \textbf{92.81} & 65.39  \\
\bottomrule
    \end{tabular}
    }
    \caption{Average accuracy across historical languages. \texttt{Lemming} and \texttt{Morfette} are shown aggregated by taking the best performing model per dataset.}
    \label{tab:hist}
\end{table}

Table~\ref{tab:hist} shows the aggregated results over all datasets in our historical language corpus.\footnote{
     We aggregate both edit-tree based approaches by selecting the best performing model for each corpus. When \texttt{Lemming} converge, the results were better than \texttt{Morfette}.
} In 4 cases (\texttt{cga}, \texttt{cgl}, \texttt{crm} and \texttt{gml}), \texttt{Lemming} failed to converge due to memory requirements exceeding 250G RAM due to the large amount of edit-trees. Following \citet{W14-1601}, we compute p-values with a Wilcoxon's signed rank test. \texttt{Sent-LM} is the best performing model with a relative improvement of 7.9\% ($p<.01$) over \texttt{Sent} and 30.72\% ($p<.01$) over the edit-tree approach on full datasets and 10.27\% ($p<.1$) and 18.66\% ($p<.01$) on ambiguous tokens. Moreover, the edit-tree approach outperforms encoder-decoder models \texttt{Plain} and \texttt{Sent} on ambiguous tokens, and it is only due to the joint loss that the encoder-decoder paradigm gains an advantage. Finally, for tokens unseen during training, the best performing model is \texttt{Sent} with a relative error reduction of 47\% ($p<.01$) over the edit-tree approach and 4.77\% ($p<.1$) over \texttt{Sent-LM}.



        

\begin{table}[htb]
    \centering
    \begin{tabular}{lccc}
\toprule

    & Full & Ambiguous & Unknown  \\

\midrule
   \texttt{K-2016}    & 91.88  & &  51.64 \\
   \texttt{Edit-tree}  & 89.01 & 91.18 & 20.46         \\
   \texttt{Plain}     & 90.21 & 87.4 & 61.93 \\
   \texttt{Sent}      & 92.55 & 91.6 & \textbf{64.61}   \\
   \texttt{Sent-LM}   & \textbf{93.25} & \textbf{93.31} & 62.1   \\
\bottomrule
        
    \end{tabular}
    \caption{Accuracy for the Gysseling subcorpora. 
}
    \label{tab:gys}
\end{table}

Table~\ref{tab:gys} compares scores for a subset from the corpora coming from the Gysseling corpus, which have been used in previous work on lemmatization of historical languages. The model described by \citet{kestemont2016lemmatization} is included as \texttt{K-2016} for comparison.\footnote{
    Unfortunately, scores on ambiguous tokens were not reported and therefore cannot be compared.
}
It is apparent that both \texttt{Sent} and \texttt{Sent-LM} outperform \texttt{K-2016} on full and unknown tokens. It is worth noting that \texttt{K-2016}, a model that uses distributed contextual features but no edit-tree induction, performs better than \texttt{Plain} -- which highlights the importance of context for the lemmatization of historical languages --, and also better than the edit-tree approaches -- which highlights the difficulty of tree induction on this dataset. We find \texttt{Sent-LM} to have a significant advantage over \texttt{Sent} on full and ambiguous tokens, but a disadvantage vs \texttt{Sent} and \texttt{Plain} on unknown tokens.

\subsection{Standard languages}\label{sec:UD}

\begin{table*}[ht]
    \centering
     \resizebox{0.9\textwidth}{!}{
    \begin{tabular}{lccccccccc}
\toprule
         & \multicolumn{3}{c}{Full} & \multicolumn{3}{c}{Ambiguous} & \multicolumn{3}{c}{Unknown}  \\
& Type 1 & Type 2 & Type 3& Type 1 & Type 2 & Type 3& Type 1 & Type 2 & Type 3\\

\midrule 
   \texttt{Edit-tree}   & 96.34 & 93.02 & \textbf{98.37} & 92.47 & \textbf{92.47} & \textbf{97.5} & 84.99 & 74.66 & \textbf{91.39} \\
   \texttt{Plain}       & 96.21 & 94.6 & 97.5  & 88.87 & 90.51 & 95.34 & 85.43 & \textbf{83.78} & 86.37 \\
   \texttt{Sent}    & 96.42 & 94.41 & 97.84 & 91.12 & 92.01 & 96.65 & \textbf{85.44} & 84.02 & 86.67 \\
   \texttt{Sent-LM} & \textbf{96.52} & \textbf{94.62} & 97.86 & \textbf{93.01} & 92.07 & 97.48 & 85.15 & 83.32 & 85.36 \\
\bottomrule
        
    \end{tabular}
    }
    \caption{Average accuracy across morphologically related standard languages. Type 1 encloses \texttt{bg}, \texttt{cs}, \texttt{lv}, \texttt{ru} and \texttt{sl}. Type 2 comprises \texttt{et}, \texttt{fi}, \texttt{hu}, \texttt{tr}. Finally, Type 3 encompasses \texttt{de}, \texttt{en}, \texttt{es}, \texttt{fr}, \texttt{it} and \texttt{nb}.}
    \label{tab:UDfam}
\end{table*}

\begin{table}[ht]
    \centering
    \begin{tabular}{lccc}
\toprule
         & Full & Ambiguous & Unknown  \\
\midrule 
   \texttt{Edit-tree}    &  96.1  & \textbf{94.35}  & \textbf{83.26}  \\
   \texttt{Plain}        &  95.93 & 91.44  & 83.02  \\
   \texttt{Sent}     &  96.19 & 93.25  & 82.61  \\
   \texttt{Sent-LM}  &  \textbf{96.28} & 94.08  & 82.58  \\
\bottomrule
    \end{tabular}
    \caption{Average accuracy across all 20 standard languages. \texttt{Lemming} and \texttt{Morfette} are shown aggregated by taking the best performing model per dataset.}
    \label{tab:UD}
\end{table}

Table~\ref{tab:UD} shows overall accuracy scores aggregated across all languages.\footnote{
    Similarly to results on historical languages, we aggregate \texttt{Morfette} and \texttt{Lemming} due to the later failing to converge on \texttt{et}. 
} We observe that on average \texttt{Sent-LM} is the best model on full datasets. However, in contrast to previous results, the edit-tree approach has an advantage over all encoder-decoder models for both ambiguous and unknown tokens. 

Since the differences in performance are not statistically significant ($p>0.05$), we seek to shed light on the advantages and disadvantages of the encoder-decoder and edit-tree paradigms by conducting a more fine-grained analysis with respect to the morphological typology of the considered languages. To this end, we group languages into morphological types depending on the dominant morphological processes of each language and aggregate scores over languages in each type:

\begin{description}[topsep=4pt]
\item[Type 1.] Balto-Slavic languages which are known for their strongly suffixing morphology and complex case system.
\item[Type 2.] Uralic and Altaic languages, which are characterized by agglutinative morphology and a tendency towards monoexponential case and vowel harmony.
\item[Type 3.] Western European languages with a tendency towards synthetic morphology and partially lacking nominal case.
\end{description}     

Table~\ref{tab:UDfam} shows accuracy scores per morphological group for each model type. It is apparent that the \texttt{Edit-tree} approach is very effective for Type 3 languages both in ambiguous and unknown tokens. In both Type 1 and Type 2 languages, the best overall performing model is  \texttt{Sent-LM}. In the case of ambiguous tokens, \texttt{Sent-LM} achieves highest accuracy for Type 1 languages, but it is surpassed by the \texttt{Edit-tree} approach on Type 2 languages. Finally, in the case of unknown tokens, we observe a similar pattern to the historical languages where \texttt{Plain} and \texttt{Sent} have an advantage over \texttt{Sent-LM}.



\section{Discussion}\label{sec:disc}

For clarity, we group the discussion of the main findings according to four major discussion points.

\paragraph{How does the joint LM loss help?} As Section~\ref{sec:results} shows, \texttt{Sent-LM} is the overall best model, and its advantage is biggest on ambiguous datasets, always outperforming the second-best encoder-decoder model on ambiguous tokens. For a more detailed comparison of the two models we tested the following two hypotheses: (i) the joint LM loss helps by providing sentence representations with stronger disambiguation capacities  (ii) The joint LM loss helps in cases when the evidence of a token-lemma relationship is sparse ---e.g in languages with highly synthetic morphological systems and in the presence of spelling variation. 

\begin{figure}[ht]
    \centering
    \includegraphics[scale=0.65]{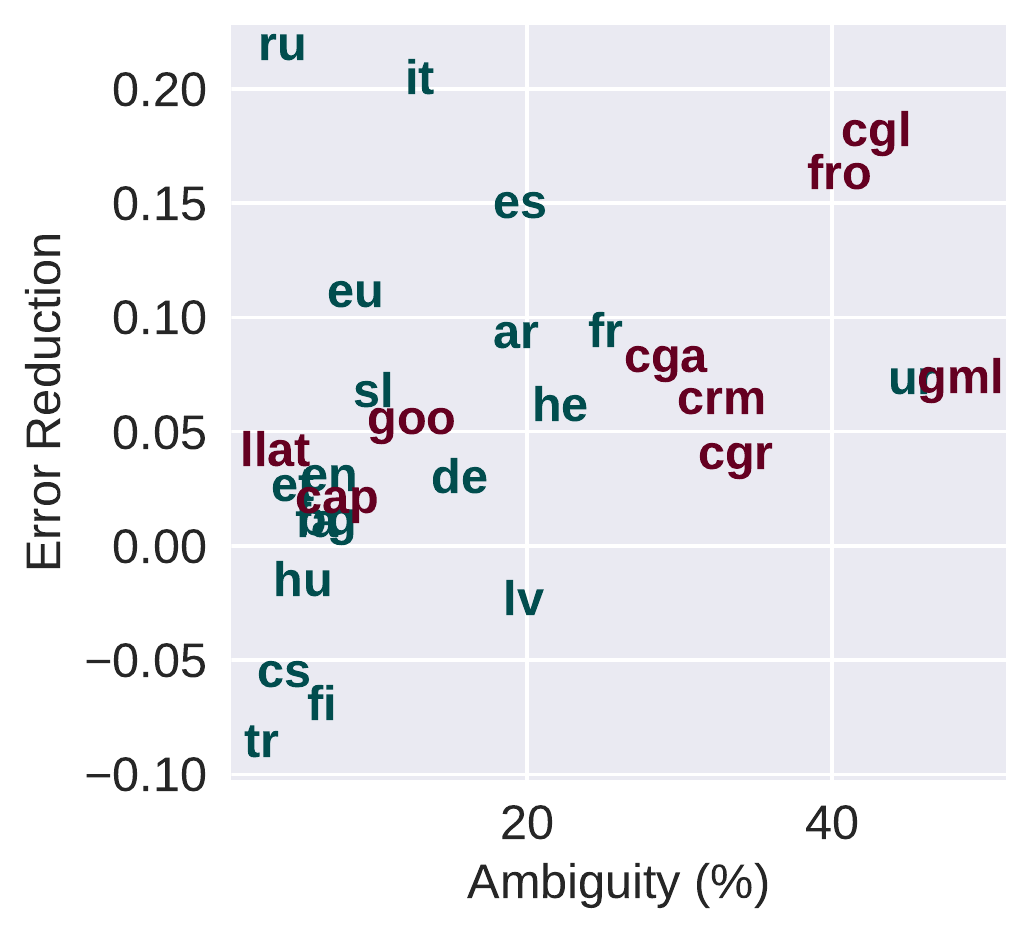}
    \caption{Error reduction of \texttt{Sent-LM} vs \texttt{Sent} by percentage of ambiguous tokens (Spearman's $R=0.53$; $p<.01$). Standard and historical languages are shown respectively in blue and red.}
    \label{fig:error-amb}
\end{figure}

\begin{figure}[ht]
    \centering
    \includegraphics[scale=0.65]{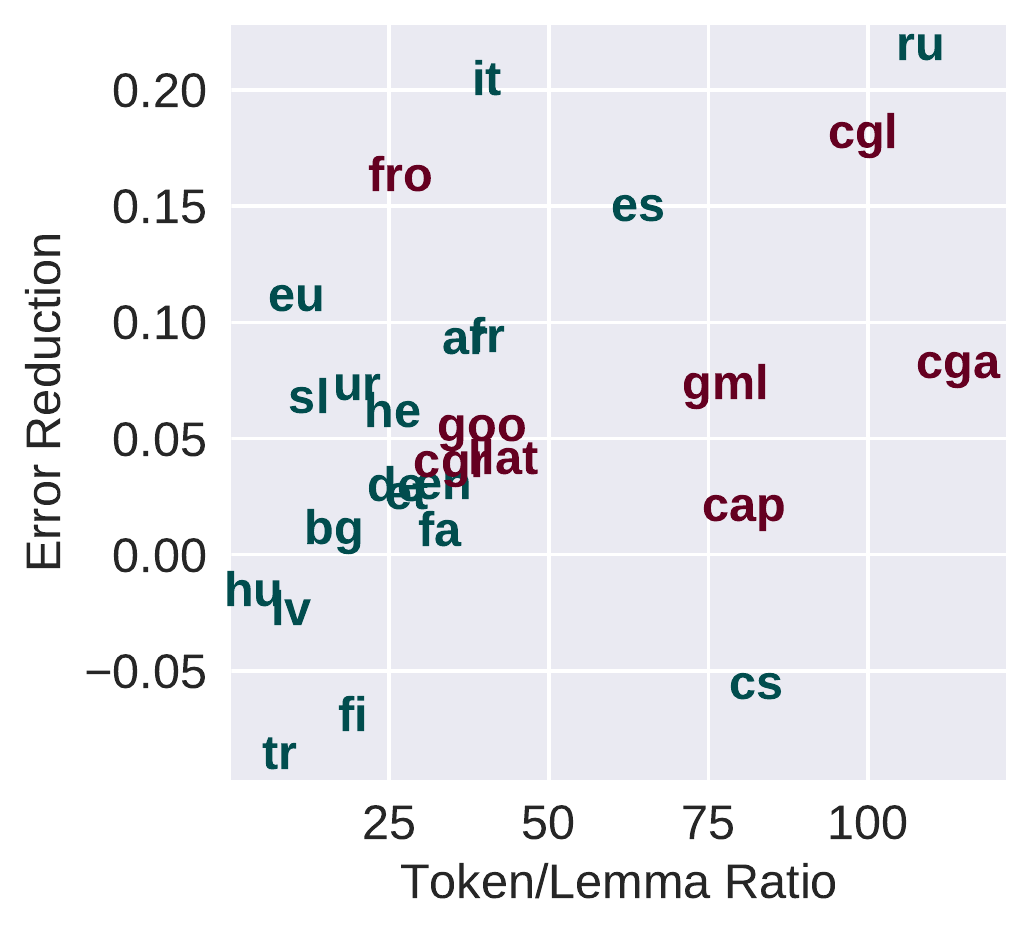}
    \caption{Error reduction of \texttt{Sent-LM} vs \texttt{Sent} by Token-Lemma ratio on 50k tokens of the training set (Spearman's $R=0.47$; $p<.05$). Standard and historical languages are shown respectively in blue and red.}
    \label{fig:error-token-lemma}
\end{figure}

As Figure~\ref{fig:error-amb} shows, improvement over \texttt{Sent} is correlated with percentage of token-lemma ambiguity in the corpus, providing evidence for hypothesis (i). Finally, as Figure~\ref{fig:error-token-lemma} shows, improvement over \texttt{Sent} is correlated with higher token-lemma ratio, suggesting that the improvement is likely to be due to learned representations that better identify the input token. These two aspects help explain the efficiency of the joint learning approach on non-standard languages where high levels of spelling variation provide increased ambiguity by conflating unrelated forms and also lower evidence for token-lemma mappings.

Another factor certainly related to the efficiency of the proposed joint LM-loss is the size of the training dataset. However, dataset size should be considered a necessary but not a sufficient condition for the feasibility of the joint LM-loss and has therefore weak explanation power for the performance of the proposed approach.

\paragraph{LM loss leads to better representations} In order to analyze the representations learned with the joint loss, we turn to ``representation probing" experiments following current approaches on interpretability \cite{Q16-1037,Adi17}. Using the same train-dev-test splits from the current study, we exploit additional POS, Number, Gender, Case and syntactic function (Dep) annotations provided in the UD corpora and compare the ability of the representations extracted by \texttt{Sent} and \texttt{Sent-LM} to predict these labels.\footnote{
    Note that not all tasks are available for all languages, due to some corpora not providing all annotations and some categories not being relevant for particular languages.
} Model parameters are frozen and a linear softmax layer $Q \in \mathbb{R}^{H\times V}$ per task is learned using a cross-entropy loss function.\footnote{
    Models trained for 50 epochs using the Adam optimizer with default learning rate and training stops after 2 epochs without accuracy increase on dev set.
}
The results of this experiment are reported in Table~\ref{tab:probe}. The classifier trained with \texttt{Sent-LM} outperforms the one with \texttt{Sent} on all considered labeling tasks, confirming the efficiency of the LM loss at extracting better representations.

\begin{table}
    \centering
    \resizebox{1\linewidth}{!}{
    
    \begin{tabular}{lccccc}
    \toprule
                  & Pos & Dep & Gender & Case & Num \\
        \midrule
        \texttt{Majority}    & 82.61 & 62.93 & 85.83 & 80.76 & 83.92 \\
        \texttt{Sent}    & 79.27 & 64.14 & 83.39 & 83.01 & 81.01 \\
        \texttt{Sent-LM} & \textbf{83.62} & \textbf{68.36} & \textbf{86.55} & \textbf{84.44} & \textbf{84.37} \\
        \midrule
        Support     & 29 & 20 & 15 & 19 & 20\\
        
    \bottomrule
    \end{tabular}
    }
    \caption{Overall accuracy of \texttt{Sent} and \texttt{Sent-LM} models and a \texttt{Majority} baseline on 5 probing tasks and actual number of languages per morphological category. All differences over \texttt{Sent} except for Case are significant at $p<0.05$. Support is shown in terms of number of languages exhibiting such grammatical distinctions.
    }
    \label{tab:probe}
\end{table}

\begin{figure}[ht]
    \centering
    \includegraphics[scale=0.65]{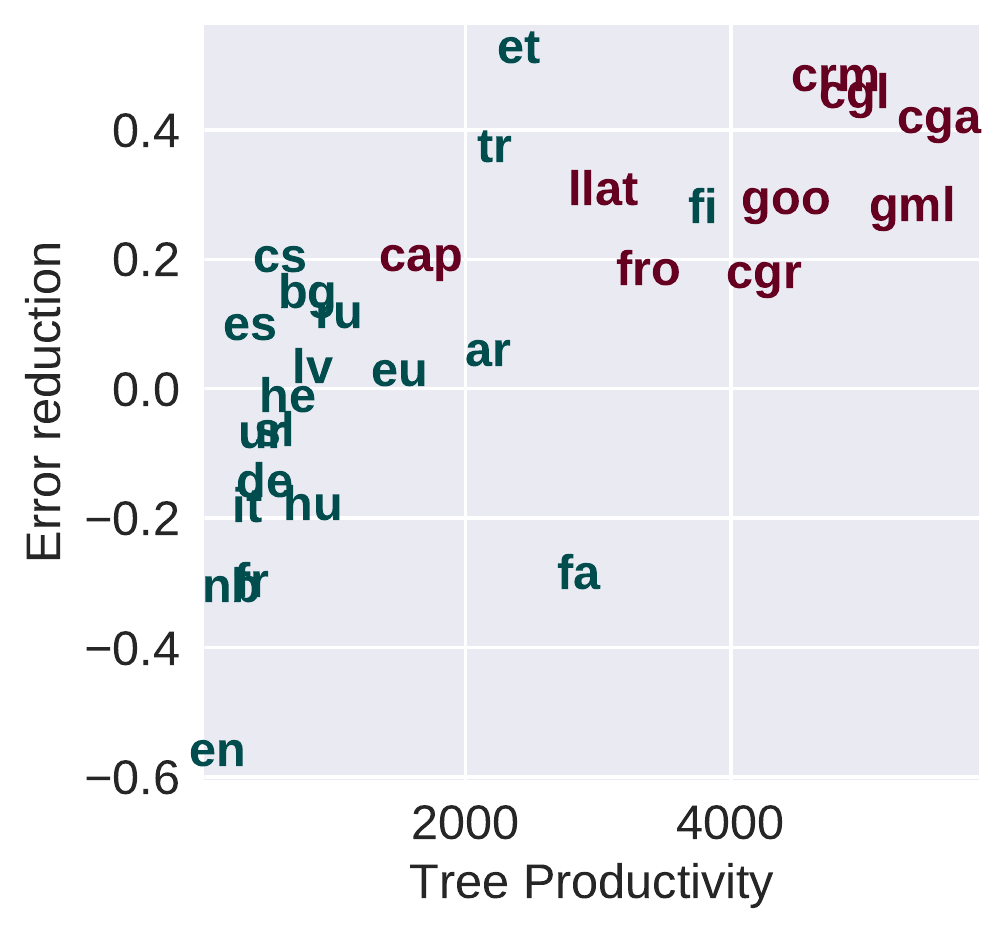}
    \caption{Error reduction of best encoder-decoder model vs best tree-edit model over tree productivity computed as number of unique binary edit-trees in the first 50k tokens of the training corpora (Spearman's $R=0.79$; $p<.001$). Standard and historical languages are shown respectively in blue and red.}
    \label{fig:error-tree}
\end{figure}

\paragraph{Edit-tree vs. Encoder-Decoder} Our fine-grained analysis suggests that the performance of the edit-tree and encoder-decoder approaches depends on the underlying morphological typology of the studied languages. Neural approaches seem to be stronger for languages with complex case systems and agglutinative morphology. In contrast, edit-tree approaches excel on more synthetic languages (e.g. Type~3) and languages with lower ambiguity (e.g. Type~2)
. Figure~\ref{fig:error-tree} illustrates that as the number of edit-trees increase the encoder-decoder models start to excel. This is most likely due to the fact that, from an edit-tree approach perspective, a large number of trees creates a large number of classes, which leads to higher class imbalance and more sparsity. However, edit-tree based approaches do outperform representation learning methods when the number of trees is low, which leads to the intuition that the edit-tree formalism does provide a useful inductive bias to the task of lemmatization and it should not be discarded in future work. Our results, in fact, point to a future direction which applies the edit-tree formalism, but alleviates the edit-tree explosion by exploiting the relationships between the edit-tree classes potentially using representation learning methods.

\paragraph{Accuracy on unknown tokens} We observe that while overall the joint loss outperforms the simpler encoder-decoder, it seems, however, detrimental to the accuracy on unknown tokens. This discrepancy is probably due to the fact that (i) unknown tokens are likely unambiguous and therefore less likely to profit from improved context representations and to (ii) our design choice of word-level language modeling, where the model is forced to predict UNK for unknown words. As \texttt{Sent-LM} is the overall best  model, in future work we will explore character-level language modeling in order to harness the full potential of the joint-training approach even on unknown tokens.

\section{Conclusion}

We have presented a method to improve lemmatization with encoder-decoder models by improving context representations with a joint bidirectional language modeling loss. Our method sets a new state-of-the-art for lemmatization of historical languages and is competitive on standard languages. Our examination of the learned representations indicates that the LM loss helps enriching sentence representations with features that capture morphological information. In view of a typologically informed comparison of encoder-decoder and edit-tree based approaches, we have shown that the latter can be very effective for highly synthetic languages. Such result might have been overlooked in previous studies due to only considering a reduced number of languages \cite{P17-1136} or pooling results across typology \cite{Bergmanis2018ContextLematus}. With respect to languages with higher ambiguity and token-lemma ratio, the encoder-decoder approach is preferable and the joint loss generally provides a substantial improvement. Finally, while other models use morphological information to improve the representation of context (e.g. edit-tree approaches), our joint language modeling loss does not rely on any additional annotation, which can be crucial in low resource and non-standard situations where annotation is costly and often not trivial.



\section{Acknowledgments}

We thank NVIDIA for donating 1 GPU that was used for the experiments in the present paper. We would also like to thank the anonymous reviewers for their valuable comments.

\bibliographystyle{acl_natbib}
\bibliography{library,2ndbib}

\newpage

\appendix
\section*{Appendix}
\section{Dataset Statistics}

Table~\ref{tab:histlang} shows the dataset sources and codes from our Historical Languages. 

\begin{table}[H]
    \centering
    \begin{tabular}{ccc}
    \toprule
        Language          & Dataset              & Code \\
        \midrule
        Medieval Latin    & Capitularia          & cap \\
                          & LLCT1                & llat \\
        Middle Dutch      & Gys (Admin)    & cga \\
                          & Gys (Literary) & cgl \\
                          & Religious & cgr \\
                          & Adelheid & crm \\
        Medieval French   & Geste                & fro \\
        Middle Low German & ReN & gml \\
        Slovenian         & goo300k              & goo \\
        \bottomrule
    \end{tabular}
    \caption{Corpus identifier and description in the historical languages dataset. ``Gys" refers to the Gysseling corpus, which consists of several subsets.}
    \label{tab:histlang}
\end{table}

Table~\ref{tab:UDlang} shows the languages from the UD corpus that were sampled for the study. We have used ISO 639-1 codes (instead of the more general ISO 639-2) in order to avoid clutter in the plots.

\begin{table}[H]
    \centering
    \makebox[1\linewidth]{
    \begin{tabular}{ccc}
    \toprule
        Language & Dataset & Code \\
        \midrule
        Arabic & Arabic-PDAT & ar \\
        Bulgarian & Bulgarian-BTB & bg \\
        Czech & Czech-CAC & cs \\
        German & German-GSD & de \\
        English & English-EWT & en \\
        Spanish & Spanish-AnCora & es \\
        Estonian & Estonian-EDT & et \\
        Basque & Basque-BDT  & eu \\
        Persian & Persian-Seraji & fa \\
        Finnish & Finnish-TDT & fi \\
        French & French-GSD & fr \\
        Hebrew & Hebrew-HTB & he \\
        Hungarian & Hungarian-Szeged & hu \\
        Italian & Italian-ISDT & it \\
        Latvian & Latvian-LVTB & lv \\
        Norwegian (Bokmaal) & Norwegian-Bokmaal & nb \\
        Russian & Russian-SynTagRus & ru \\
        Slovenian & Slovenian-SSJ & sl \\
        Turkish & Turkish-IMST & tr \\
        Urdu & Urdu-UDTB & ur \\ 
        \bottomrule
    \end{tabular}
    }
    \caption{Standard language datasets from the Universal Dependencies (v2.2) corpus.}
    \label{tab:UDlang}
\end{table}

\end{document}